\pdfoutput=1

\documentclass[11pt]{article}

\usepackage[final]{acl}

\usepackage{times}
\usepackage{latexsym}

\usepackage[T1]{fontenc}

\usepackage[utf8]{inputenc}

\usepackage{microtype}

\usepackage{inconsolata}

\usepackage{graphicx}

\usepackage{amsmath}
\usepackage{amsfonts}
\usepackage{amssymb}
\usepackage{booktabs}
\usepackage{graphicx}
\usepackage{cleveref}
\usepackage{enumitem}
\usepackage{xcolor}
\usepackage{xspace}
\usepackage{multirow}

\def\eg{{\em e.g.,}\xspace}
\def\ie{{\em i.e.,}\xspace}
\crefname{section}{\S}{\S\S}
\Crefname{section}{\S}{\S\S}
\crefformat{section}{\S#2#1#3}
\crefname{table}{Table}{Tables}
\crefname{figure}{Figure}{Figures}
\crefname{appendix}{Appendix}{Appendix}

\makeatletter
\DeclareRobustCommand*{\escapeus}[1]{%
  \begingroup\@activeus\scantokens{#1\endinput}\endgroup}
\begingroup\lccode`\~=`\_\relax
   \lowercase{\endgroup\def\@activeus{\catcode`\_=\active \let~\_}}
\makeatother
\usepackage[scaled=0.86]{helvet}
\newcommand{\makesf}[1]{\textsf{{\escapeus{#1}}}}

\def\popqa{\makesf{PopQA}\xspace}
\def\triviaqa{\makesf{TriviaQA}\xspace}
\def\nqopen{\makesf{NQ-Open}\xspace}
\def\gsm{\makesf{GSM8K}\xspace}
\DeclareMathOperator*{\expect}{\mathbb{E}}

\title{Self-Training Large Language Models for Tool-Use \\ Without Demonstrations}

\author{
\normalsize
Ne Luo$^1$\qquad
Aryo Pradipta Gema$^1$\qquad
Xuanli He$^2$\qquad
Emile van Krieken$^1$\\ \bf
\normalsize
Pietro Lesci$^3$\qquad
Pasquale Minervini$^{1,4}$\\
\normalsize
$^1$University of Edinburgh \quad
$^2$University College London \quad
\normalsize
$^3$University of Cambridge \quad
$^4$Miniml.AI\\
\normalsize
\href{mailto:n.luo-5@sms.ed.ac.uk}{\makesf{n.luo-5@sms.ed.ac.uk}} \quad \href{mailto:p.minervini@ed.ac.uk}{\makesf{p.minervini@ed.ac.uk}}\\
}

\begin{document}

\maketitle

\begin{abstract}
Large language models (LLMs) remain prone to factual inaccuracies and computational errors, including hallucinations and mistakes in mathematical reasoning. 
Recent work augmented LLMs with tools to mitigate these shortcomings, but often requires curated gold tool-use demonstrations.
In this paper, we investigate whether LLMs can learn to use tools \emph{without} demonstrations.
First, we analyse zero-shot prompting strategies to guide LLMs in tool utilisation. 
Second, we propose a self-training method to synthesise tool-use traces using the LLM itself.
We compare supervised fine-tuning and preference fine-tuning techniques for fine-tuning the model on datasets constructed using existing Question Answering (QA) datasets, \ie \triviaqa and \gsm.
Experiments show that tool-use enhances performance on a long-tail knowledge task: 3.7\% on \popqa, which is used solely for evaluation, but leads to mixed results on other datasets, \ie \triviaqa, \gsm, and \nqopen. 
Our findings highlight the potential and challenges of integrating external tools into LLMs without demonstrations.\footnote{Code available at \href{https://github.com/neneluo/llm-tool-use}{\makesf{github.com/neneluo/llm-tool-use}}.}
\end{abstract}

\section{Introduction}
Large language models (LLMs) have shown state-of-the-art performance in many natural language processing tasks and demonstrated ``emergent abilities'':  the capability to perform new tasks without updating their parameters via gradient descent 
\citep{DBLP:conf/nips/BrownMRSKDNSSAA20, DBLP:journals/corr/abs-2211-05100, DBLP:journals/jmlr/ChowdheryNDBMRBCSGSSTMRBTSPRDHPBAI23, DBLP:journals/corr/abs-2302-13971}. Specifically, by simply being provided with task instructions, sometimes supplemented with a few examples, LLMs can achieve comparable performance to fine-tuning-based methods \citep{DBLP:conf/nips/BrownMRSKDNSSAA20, DBLP:conf/iclr/WeiBZGYLDDL22, DBLP:conf/iclr/AkyurekSA0Z23, DBLP:conf/icml/OswaldNRSMZV23}.
Despite the remarkable performance, LLMs may still generate inaccurate or unfactual texts, \ie hallucinations \citep{DBLP:conf/nips/LewisPPPKGKLYR020, DBLP:journals/corr/abs-2404-05904} or perform incorrect computations \citep{DBLP:conf/icml/GaoMZ00YCN23} if they solely rely on their internal parametric knowledge.
Motivated by these shortcomings, we explore approaches that augment LLMs with external tools \citep{DBLP:journals/tmlr/MialonDLNPRRSDC23}---such as a calculator or a search engine---to enhance their reasoning and problem-solving abilities. 

\begin{figure}[t]
    \centering
    \includegraphics[width=1\columnwidth]{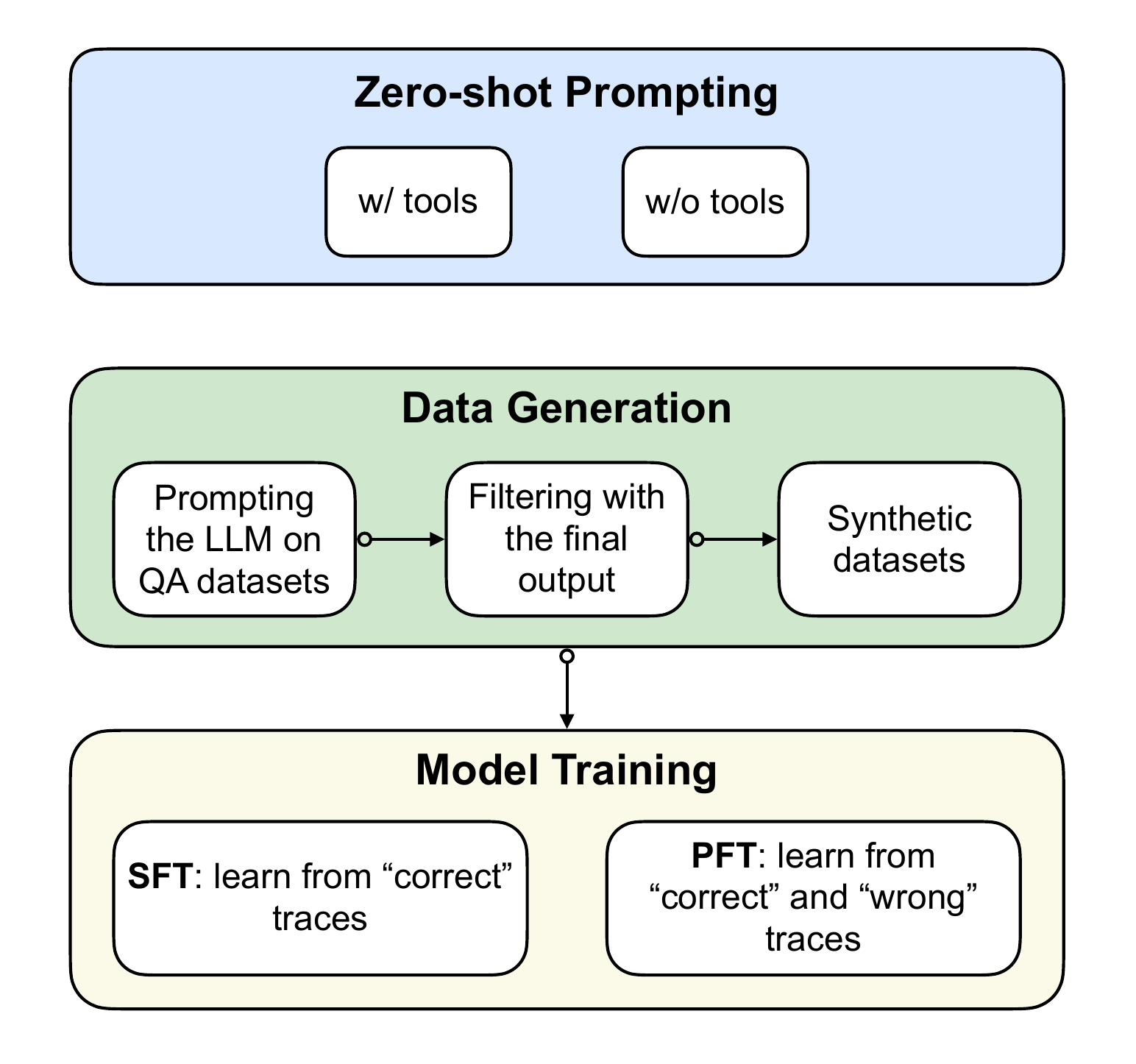}
    \caption{The overall workflow of our work. SFT: supervised fine-tuning; PFT: preference fine-tuning.}
    \label{fig:workflow}
\end{figure}

Augmenting LLMs with tools has become an active research area in recent years \citep{DBLP:journals/tmlr/MialonDLNPRRSDC23, DBLP:conf/nips/SchickDDRLHZCS23, DBLP:conf/iclr/QinLYZYLLCTQZHT24, DBLP:conf/iclr/0040CW0T0024}. Current methods primarily follow two approaches \citep{DBLP:journals/corr/abs-2403-15452}: (i) \textbf{prompting}, which leverages the in-context learning ability of large-scale models \citep{DBLP:conf/iclr/YaoZYDSN023, DBLP:conf/nips/LuPCGCWZG23}; and (ii) \textbf{fine-tuning}, with a primary focus on Supervised Fine-Tuning (SFT), which trains LLMs on datasets of tool-use examples. These datasets are typically sampled from large-scale data \citep{DBLP:conf/nips/SchickDDRLHZCS23} or generated from more advanced LLMs, such as ChatGPT~\citep{DBLP:conf/nips/YangSLZGLS23, DBLP:conf/iclr/QinLYZYLLCTQZHT24, DBLP:conf/iclr/0040CW0T0024}. While these approaches are highly effective and demonstrate impressive performance, they are resource-intensive and challenging to generalise.

Additionally, SFT with tool-use datasets alone can be suboptimal, as tool-use for LLMs is an inherently open-ended task: there is no oracle tool-use trace that specifies a single, unique solution to a given problem with specific tools. Many tools share overlapping functionality, and each tool can be used in multiple ways. Also, the helpfulness of tools depends on the LLM itself \citep{DBLP:conf/nips/SchickDDRLHZCS23}. This led us to consider optimising the model in a way that better captures the open-ended feature of tool-use, where Preference Fine-Tuning (PFT) generally helps. While concurrent work applied PFT to augment the LLMs' ability in specific domains, \eg as math agents \citep{xiong2024buildingmathagentsmultiturn, DBLP:conf/acl/WangLL24}, its potential in general tool-use scenario remains under-explored.

Inspired by these challenges, we explore whether LLMs can learn tool-use without demonstrations, leveraging different optimisation objectives; that is, we answer the question: \emph{can we teach LLMs to use tools without oracle tool-use traces?} 

First, we study zero-shot prompting approaches that utilise LLMs' instruction-following ability. This serves as a baseline for the tool-use performance of LLMs without performing gradient updates. Second, we propose a self-training approach to synthesise datasets containing tool-use traces via the LLM itself, which could be used for model fine-tuning. To improve data quality, we further explore employing different filtering strategies based on the final output and additional criteria. Then, we explore different fine-tuning objectives: SFT and PFT. \cref{fig:workflow} shows an overview of our work.

Our experimental results across multiple Question Answering (QA) datasets, including \triviaqa, \gsm, \nqopen and \popqa, show that zero-shot prompting alone enables LLMs to use tools to some extent but may lead to degraded performance, depending on the model scale and tasks. The proposed self-training method for tool-use, which trains the model on the synthetic datasets generated from \triviaqa and \gsm, improves model performance on a long-tail knowledge task \popqa but yields mixed performance on other datasets. These findings suggest that the LLMs learn tool-use even without explicit demonstrations, but the performance gain is mainly shown when the model's knowledge may be insufficient to solve a task, while inappropriate tool-use can introduce additional challenges.

\section{Background}
\label{sec:background}

In this section, we review two common methods used to fine-tune LLMs.

\paragraph{Supervised Fine-Tuning.}
SFT is performed after the pre-training of LLMs and generally enhances the model performance in specific downstream tasks, such as summarization, QA, etc \citep{DBLP:conf/nips/BrownMRSKDNSSAA20}. Instruction fine-tuning is a special form of SFT that aims to optimise the LLMs to follow human instructions, treating instruction following as a type of downstream task \citep{DBLP:conf/iclr/WeiBZGYLDDL22}. Given an instruction fine-tuning dataset, such as SUP-NATINST \citep{DBLP:conf/emnlp/WangMAKMNADASPK22}, the training of the LLM under SFT optimises model parameters $\theta$ by minimising the negative log-likelihood loss $\mathcal{L}_{\text{SFT}}$, defined as follows:
\begin{equation} \label{eq:lsft}
    \mathcal{L}_{\text{SFT}} = -\sum_{i=1}^{n} \log P(w_i|s, w_{0}, \cdots, w_{i-1}; \theta),
\end{equation}
where $s$ is an instruction, $w_i$ is the $i$-th token in the response, and $n$ is the response length.

\paragraph{Preference Fine-Tuning.}
PFT is a post-training technique used in addition to SFT, aligning model responses with human preferences by fine-tuning LLMs on pairwise preference data. The method is initially introduced as Reinforcement Learning from Human Feedback~\citep[RLHF;][]{DBLP:conf/nips/Ouyang0JAWMZASR22}.
Given an instruction, preference data are collected by annotating the human preference between two responses. Instead of strictly optimising LLMs to follow specific responses, LLMs are optimised to adhere to a policy of favouring human-preferred responses. In this way, LLMs learn to better capture the open-ended characteristics of conversations and better align with human values. 

\citet{DBLP:conf/nips/RafailovSMMEF23} introduce Direct Preference Optimization (DPO), an RL-free algorithm that directly optimises the policy $\pi_{\theta}$ of LLMs by implicitly adjusting the likelihood of preferred and dispreferred responses.
The DPO loss is derived by directly approximating the optimal policy according to the preference data, significantly simplifying the RLHF pipeline. Given the preference dataset $\mathcal{D}$, instruction $x$, preferred response $y_\text{p}$ and dispreferred response $y_\text{d}$, the DPO loss is defined as:
\begin{align} \label{eq:dpo}
    \mathcal{L}_{\makesf{DPO}} = 
    &- \expect_{(x, y_\text{p}, y_\text{d}) \sim \mathcal{D}} \Bigg[ \log \sigma \Bigg( \beta \log \frac{\pi_{\theta}(y_\text{p} \mid x)}{\pi_{\makesf{REF}}(y_\text{p} \mid x)} \nonumber\\ 
    &- \beta \log \frac{\pi_{\theta}(y_\text{d} \mid x)}{\pi_{\makesf{REF}}(y_\text{d} \mid x)} \Bigg) \Bigg]
\end{align}
where $\sigma$ is the sigmoid function, $\beta$ controls the divergence between the policy $\pi_{\theta}$ to be optimised and the reference policy $\pi_{\makesf{REF}}$, a lower value indicates higher divergence.

\section{Approach}
In this section, we introduce the tools used to augment LLMs, and describe the procedure for generating tool-use datasets for model fine-tuning.

\subsection{Tools}

Inspired by the tool-use setting in \citep{DBLP:conf/nips/SchickDDRLHZCS23, DBLP:conf/iclr/0040CW0T0024}, we developed a set of tools that can be used by LLMs, including a calculator, a Wikipedia search engine, and a machine translator. The tools serve the purpose of aiding LLMs in various areas, such as mathematical calculation, real-world information retrieval, and low-resource language understanding.

\paragraph{Calculator.} 
The calculator assists LLMs by producing accurate mathematical calculation results. It supports basic operations between numbers like addition, subtraction, etc. Given a mathematical formula, the output is the computation result.

\paragraph{Wikipedia Search Engine.}
The Wikipedia search engine assists LLMs in searching for relevant information from Wikipedia documents. The Wikipedia search engine was implemented with a BM25-based information retrieval model. Given a query, the information retrieval model retrieves the most relevant Wikipedia documents.\footnote{The documents are from a pre-built Wikipedia dump \makesf{wikipedia-kilt-doc} with index version \makesf{lucene-index.wikipedia-kilt-doc.20210421.f29307.tar.gz}.}

\paragraph{Machine Translator.}
The machine translator is defined as the setting of translating low-resource languages into English, which can potentially aid LLMs in understanding low-resource languages. The machine translator was implemented with the open-source multilingual machine translation model No Language Left Behind~\cite[NLLB;][]{DBLP:journals/corr/abs-2207-04672}, which supports 200 languages. We used a distilled version\footnote{\href{https://huggingface.co/facebook/nllb-200-distilled-600M}{\makesf{huggingface.co/facebook/nllb-200-distilled-600M}}.} for computational efficiency. Given a query, the machine translator outputs its corresponding English translation.

\subsection{Synthesising Tool-use Dataset}
Given an instruction fine-tuned LLM\footnote{We used the Llama-3-8B-Instruct model as the LLM to synthesise data in the following experiments. The model is available at: \href{https://huggingface.co/meta-llama/Meta-Llama-3-8B-Instruct}{\makesf{huggingface.co/meta-llama/Meta-Llama-3-8B-Instruct}}.}, we generated synthetic tool-use datasets via the model itself through the following steps:
\begin{enumerate}[leftmargin=*]
    \item \textbf{Tools collection:} First, we created a tool pool by defining a set of functions that the model could utilise as tools. For each tool, we provided a concise usage description, showing the type of problem the tool can help with, as well as its input and output format.
    \item \textbf{QA datasets collection:} Second, we collected some QA datasets from existing NLP datasets, \eg \gsm, that were likely to benefit from tool integration. These datasets include questions that external tools, \eg calculator, could potentially improve the model's ability to produce accurate answers.
    \item \textbf{Data generation:} Third, we generated the synthetic dataset with the instruction fine-tuned LLM. Specifically, for each question from the QA dataset, we prompted the model with instructions that describe the available tools, encouraging the model to provide an answer that potentially utilises tools.
    \item \textbf{Data filtering:} Then, we designed a data filtering process to ensure data quality. In this step, we used the correctness of the answer given a question to serve as a proxy for identifying ``correct'' tool-use traces, similar to \citet{DBLP:conf/nips/ZelikmanWMG22}. If the model provided a correct final answer, we inferred the solution path is valid. The specific data filtering strategies for SFT and PFT are described in \cref{subsec:learning}.
\end{enumerate}

\subsection{Learning to Use Tools}
\label{subsec:learning}

We describe the prompting approach we use to enable zero-shot tool-use and outline how we create synthetic datasets for fine-tuning LLMs through SFT and PFT. 

\paragraph{Prompting.} 
As the instruction fine-tuned LLMs were already optimised to follow human instructions, we hypothesised that the LLM could obtain the tool-use ability to some extent by solely learning from instructions, \ie prompts. We used the following prompt format: a short QA task description, an expected response format, and a brief description of the tools' applicable domain. The task description suggests the goal of the task is to answer questions briefly. \
We employed zero-shot prompting without providing any tool-use examples and adopted the zero-shot Chain-of-Thought (CoT) method \citep{DBLP:conf/nips/KojimaGRMI22}. The LLMs' response format design was inspired and adapted from \citep{DBLP:conf/iclr/YaoZYDSN023, DBLP:conf/iclr/0040CW0T0024}. \cref{tab:prompt_tool_single_step} provides an example of the tool-use prompt that enables the LLM to perform single-step tool-use, either to request a single tool or multiple tools simultaneously. Variants of the prompt used for ablation studies are shown in \cref{sec:prompts}. 

\begin{table}[!t]
    \centering
    \footnotesize
    \begin{tabular}{p{0.9\columnwidth}}
        \toprule
        You are an advanced AI agent designed to answer questions. You can use your own knowledge to answer the questions, or use external tools to gather information before answering. However, you can only request the use of tools once. Answer in a few words. Let's think step by step. \\ \\
        Respond in the following format: \\
        Thought: decide whether to answer using your own knowledge or utilise external tools. \\
        Action: specify the tool here using the format `ToolName[query]' if you decide to use tools. \\
        Rationale: justify your answer by providing intermediate reasoning steps for your answer, based either on your own knowledge or the received tool responses. \\
        Answer: (1) if using your own knowledge, provide your answer here; (2) if using tools, leave this part empty until the tool's response is received. \\ \\
        Below are the external tools you can use: \\
        1. Calculator[query]: this tool helps you perform simple mathematical computations with real numbers. Use the formula as the input query, the tool response will be the result.  \\
        2. WikipediaSearch[query]: this tool helps you search for information from Wikipedia. Use a short keyword as the input query, the tool response will be the corresponding information. \\
        3. MachineTranslator[query]: this tool helps you understand low-resource languages by translating them to English. Use the sentence you want to translate as the input query, the tool response will be the translation. \\
        \bottomrule
    \end{tabular}
    \caption{System prompt for single-step tool-use.}
    \label{tab:prompt_tool_single_step}
\end{table}

\paragraph{Supervised Fine-Tuning.}
For SFT, we experimented with two training data filtering strategies under the same data generation procedure (prompting with \cref{tab:prompt_tool_single_step}): (i) SFT (tools data): the training data contains the instances from the training set of QA datasets where the model uses tools and provides a correct answer;
and (ii) SFT (mixture data): the training data contains the instances where the model answers correctly regardless of whether or not the tool is used during the process.
After data filtering, we re-constructed the full conversation history of the filtered cases to be the training data. 

\paragraph{Preference Fine-Tuning.} 
The triplet format data are required to form the PFT dataset: a prompt (the question), a preferred response and a dispreferred response. To generate this dataset, we conducted system inference on the training sets of QA datasets under two conditions: with and without the use of tools. We then filtered the generated data based on the following criteria: (i) the model provided the correct answer with access to tools (using the prompt in \cref{tab:prompt_tool_single_step}); (ii) the model provided the wrong answer for the same question without access to tools (using the prompt in \cref{tab:prompt_no_tool_zero_shot}). 
Then, we experimented with fine-tuning the model on tool-use data with DPO under two data format settings: (i) DPO (conversation): the preferred and dispreferred data encompass the entire conversation following the question; (ii) DPO (response): the preferred and dispreferred data are the model's single response following the question.

\section{Experimental Setup}

In this section, we describe the datasets we use for experiments and detail our research questions.

\subsection{Datasets}
To effectively benchmark the tool-use LLMs, we employed datasets from two categories of tasks that could potentially benefit from tool integration. The detailed dataset statistics are shown in \cref{sec:dataset_statistic}.

\paragraph{Open-domain QA.} 
We experimented with three open-domain QA datasets: \triviaqa \citep{DBLP:conf/acl/JoshiCWZ17}, Natural Question-Open~\citep[\nqopen;][]{DBLP:conf/acl/LeeCT19} and \popqa~\citep{DBLP:conf/acl/MallenAZDKH23}. Questions in \triviaqa require trivia knowledge. \nqopen contains real-world questions asked by actual Google Search users. \popqa consists of questions requiring knowledge of long-tail Wikipedia entities. Thus, these datasets can potentially benefit from the use of knowledge from external tools, \eg Wikipedia search engine and machine translator. We used \nqopen and \popqa solely for model evaluation. For model training, we randomly sampled a subset from the original training set for \triviaqa to match the scale of another dataset category. Similarly, we randomly selected portions from the original validation sets or test sets to be validation sets and test sets. 

\paragraph{Mathematical Reasoning.}
We experimented with the Grade School Math dataset~\citep[\gsm;][]{DBLP:journals/corr/abs-2110-14168}, a QA dataset composed of grade school math word problems, which could potentially benefit from the usage of a calculator. For our experiments, we used the original training set for model training and randomly split the original test set into the validation and test set in approximately equal proportions.

\subsection{Experiments}

In the following section, we break the main research question into three sub-research questions and conduct corresponding experiments:
\begin{itemize}[leftmargin=*]
    \item \textbf{RQ1 (\cref{sec:exp-prompt}):} Can we instruct LLMs to use tools without training them? 
    \item \textbf{RQ2 (\cref{sec:exp-sft}):} Can we further improve LLMs' tool-use ability by conducting SFT in a self-training manner?
    \item \textbf{RQ3 (\cref{sec:exp-pft}):} Can we use PFT to teach LLMs to use tools? 
\end{itemize}
All experiments were conducted on the Llama 3 instruction fine-tuned models \citep{DBLP:journals/corr/abs-2407-21783}. The model fine-tuning details on the synthetic datasets are shown in \cref{sec:training_details}. Inference details are shown in \cref{sec:interaction}. For evaluation, we employed two sets of automatic evaluation metrics to evaluate both the tool-use ability and generation quality of LLMs, which are defined in \cref{sec:metrics}.

\section{Results and Analysis}

In this section, we present the experimental results and analysis of approaches for teaching LLMs to use tools without demonstrations.

\subsection{Prompting}
\label{sec:exp-prompt}
The experimental results in \cref{tab:prompting_acc} and \cref{tab:prompting_rate} showed an initial effort to instruct LLMs to use tools by prompting the instruction fine-tuned Llama 3 models with different instructions. The detailed prompts are shown in \cref{sec:prompts}. 

\paragraph{Results on Llama-3-8B-Instruct.} 
When we prompted the LLM to answer questions with its own knowledge, zero-shot CoT prompting showed an improvement compared to the prompt of not using CoT, especially in the math dataset, \ie \gsm, where the model benefited greatly from using explicit intermediate reasoning steps. Under the tool-use setting, results in \cref{tab:prompting_rate} show that the LLM had a positive invoke rate and pass rate for tools, indicating that the LLM was aware of the existence of tools and knew how to use them to some extent. \cref{tab:use_tool_right} shows an example that the model called the calculator correctly and yielded the right answer after getting responses from the tool, although we had not fine-tuned the LLM on tool-use datasets yet. However, we observed a performance drop in \cref{tab:prompting_acc} when we allowed the model to use tools in both single-step and multi-step tool-use scenarios. This phenomenon indicates that the model can be harmed if it uses tools inappropriately. We provide extended analysis on the results of single-step tool-use in \cref{sec:analysis}.

\begin{table}[!t]
\centering
\small
\resizebox{\columnwidth}{!}{
\begin{tabular}{lccccc}
\toprule
\multirow{2}{*}{\textbf{Prompt}}                  & \multirow{2}{*}{\textbf{Size}}& \multicolumn{2}{c}{\textbf{\triviaqa}} & \multicolumn{2}{c}{\textbf{\gsm}} \\
\cmidrule(lr){3-4}\cmidrule(lr){5-6}
                        &     & \textbf{EM}      & \textbf{Acc}            & \textbf{EM}      & \textbf{Acc}      \\
\midrule
No tool                 & 8B  & 62.6	& 73.6           & 1.4     & 36.5 \\
No tool + CoT           & 8B  & 52.8	& \textbf{77.9}  & 8.6     & \textbf{66.9} \\
Tools + Single-step     & 8B  & 56.9 	& 75.8           & 22.0    & 64.2 \\
- w/o Rationale         & 8B  & 35.7 	& 72.1           & 2.9	   & 56.6 \\
Tools + Multi-step      & 8B  & 27.7	& 54.8	         & 19.1	   & 53.1 \\
\midrule
No tool                 & 70B & 70.4    & 87.8           & 3.1     & 66.0 \\
No tool + CoT           & 70B & 79.6    & \textbf{88.8}  & 21.2    & 45.8 \\
Tools + Single-step     & 70B & 57.6    & 77.7           & 51.8    & \textbf{75.1} \\
\bottomrule
\end{tabular}
}
\caption{Experimental results of Llama 3 instruction fine-tuned models on the validation sets in zero-shot setting given different prompts. EM: exact match; Acc: accuracy. The detailed prompts are shown in \cref{sec:prompts}. All numbers shown in the table are in percentages.}
\label{tab:prompting_acc}
\end{table}

\begin{table}[!t]
\centering
\small
\setlength{\tabcolsep}{2.5pt}
\resizebox{\columnwidth}{!}{
\begin{tabular}{lccccccc}
\toprule
\multirow{2}{*}{\textbf{Prompt}}                 & \multirow{2}{*}{\textbf{Size}} & \multicolumn{3}{c}{\textbf{\triviaqa}} & \multicolumn{3}{c}{\textbf{\gsm}} \\
\cmidrule(lr){3-5}\cmidrule(lr){6-8}
                       &     & \textbf{IR}    & \textbf{PR}    & \textbf{AR}      & \textbf{IR}      & \textbf{PR}    & \textbf{AR}      \\
\midrule
Tools + Single-step    & 8B  & 14.5	 & 99.3	 & 55.2    & 7.2	 & 72.3	 & 12.8 \\
- w/o Rationale        & 8B  & 29.2	 & 99.7  & 55.5    & 13.5	 & 75.0  & 13.6 \\
Tools + Multi-step     & 8B  & 78.3  & 99.4	 & 63.9    & 55.7    & 72.9	 & 16.6	\\
\midrule
Tools + Single-step    & 70B & 48.5  & 99.0  & 63.1    & 53.8    & 70.6  & 36.9 \\
\bottomrule
\end{tabular}
}
\caption{Experimental results of Llama 3 instruction fine-tuned models on the validation sets in zero-shot setting given different prompts. IR: invoke rate; PR: pass rate; AR: answerable rate. The detailed prompts are shown in \cref{sec:prompts}. All numbers shown in the table are in percentages.}
\label{tab:prompting_rate}
\end{table}

\begin{table}[!t]
    \centering
    \definecolor{DarkerGreen}{RGB}{34,139,34}
    \footnotesize
    \begin{tabular}{p{0.9\columnwidth}}
        \toprule
        \underline{\textbf{\textsc{Non-tool-use example}}} \\
        \textbf{User:} \\
        Hannah slips on a banana peel and breaks her arm. The doctor charges her \$200 for the cast, \$300/hour for a 30-minute visit, \$4/pill for 30 painkillers, and \$6/hour for a mandatory minimum of 2 hours of parking. How much does the doctor's visit cost total? \\
        \textbf{Assistant:} \\
        Thought: Let's break down the costs into separate components and calculate the total. \\
        Answer: \$200 (cast) + \$45 (30-minute visit) + \$120 (30 painkillers) + \$12 (2 hours of parking) = \textcolor{red}{\$477. (wrong answer)}\\
        \midrule
        \underline{\textbf{\textsc{tool-use example}}} \\
        \textbf{User:} \\
        Hannah slips on a banana peel and breaks her arm. The doctor charges her \$200 for the cast, \$300/hour for a 30-minute visit, \$4/pill for 30 painkillers, and \$6/hour for a mandatory minimum of 2 hours of parking. How much does the doctor's visit cost total? \\
        \textbf{Assistant:} \\
        Thought: I will use external tools to help me with the calculation. \\
        Action: Calculator[(30*300)/60 + 200 + (30*4) + (6*2)] \\
        Rationale: To calculate the total cost, I will first calculate the doctor's visit fee, which is 30 minutes * \$300/hour = \$150. Then, I will add the cost of the cast (\$200), the painkillers (\$4/pill * 30 = \$120), and the parking fee (\$6/hour * 2 hours = \$12). \\
        Answer: \\
        \textbf{User:} \\
        Response from tool Calculator[(30*300)/60 + 200 + (30*4) + (6*2)] are: 482.0 \\
        \textbf{Assistant:} \\
        Answer: \textcolor{DarkerGreen}{482.0 (correct answer)} \\
        \bottomrule
    \end{tabular}
    \caption{An example of the model getting the right answer when using tools from the validation set of \gsm.}
    \label{tab:use_tool_right}
\end{table}

\paragraph{Results on Llama-3-70B-Instruct.} 
Results in \cref{tab:prompting_acc} show the accuracy of the 70B model with tool-use dropped on \triviaqa and increased on \gsm compared to not using tools. The invoke rates showed in \cref{tab:prompting_rate} were much higher for the 70B model than the 8B model, indicating increased tool integration in larger models. In \triviaqa, the model shows a similar pattern as in the 8B model. However, in \gsm, we notice that the 70B model showed abnormally lower accuracy compared to the 8B model in no tool CoT prompting, where the model was also asked to follow a specific answer format. This could be because the provided human-written prompts were suboptimal and had hurt the model's performance by constraining it from following the designed answer format, which aligns with prior research that constrained answer formats can negatively influence model performance \citep{tam2024letspeakfreelystudy}. Therefore, the 70B model performed worse because its better instruction-following ability makes it more sensitive to prompts. The tools appeared helpful for the 70B model in \gsm, suggesting that even with prompting-based methods alone, tools can be beneficial for large models in complex reasoning tasks.

\begin{table*}[!t]
\centering
\small
\renewcommand{\arraystretch}{0.8}
\begin{tabular}{lcccccccc}
\toprule
\multirow{2}{*}{\textbf{System}}             & \multicolumn{2}{c}{\textbf{\triviaqa}} & \multicolumn{2}{c}{\textbf{\gsm}} & \multicolumn{2}{c}{\textbf{\nqopen}} & \multicolumn{2}{c}{\textbf{\popqa}} \\
\cmidrule(lr){2-3}\cmidrule(lr){4-5}\cmidrule(lr){6-7}\cmidrule(lr){8-9}
                   & \textbf{EM}      & \textbf{Acc}            & \textbf{EM}      & \textbf{Acc}           & \textbf{EM}      & \textbf{Acc}           & \textbf{EM}   & \textbf{Acc}           \\
\midrule
Prompting (no tool) & 52.8	 & \textbf{77.9}  & 8.6	    & \textbf{66.9} & 11.6    & \textbf{40.3} & 18.0 & 31.6          \\
Prompting (tools)   & 56.9	 & 75.8           & 22.0	& 64.2          & 15.6    & 38.9          & 22.3 & \textbf{34.2} \\
\midrule
SFT (tools data)   & 35.0	 & 65.9           & 16.6	& 48.6          & 10.1    & 37.7          & 18.1 & 35.0          \\
SFT (mixture data) & 56.1	 & 75.9           & 17.5	& \textbf{64.6} & 14.0    & \textbf{40.3} & 22.4 & \textbf{35.7} \\
SFT (no tool data) & 56.8    &\textbf{ 77.6}  & 17.8    & \textbf{64.6} & 15.0    & 39.5          & 22.3 & 33.6          \\
\midrule
DPO (conversation) & 56.1	 & \textbf{74.5}  & 21.1	& 62.2          & 14.5    & 38.4          & 22.3 & \textbf{34.6} \\
SFT + DPO          & 56.8	 & 73.6           & 17.4	& \textbf{64.2} & 15.5    & \textbf{39.1} & 22.3 & 33.0          \\
\midrule
DPO (response)     & 56.4	 & \textbf{75.4}  & 18.2	& 62.9          & 14.4    & 37.8          & 22.6 & 35.1          \\
DPO $\beta=0.01$   & 54.8    & 73.3           & 19.1    & 64.0          & 14.7    & 38.0          & 23.6 & \textbf{35.6} \\
DPO $\beta=0.5$    & 53.9	 & 73.0           & 17.7	& \textbf{64.2} & 14.9    & \textbf{38.7} & 23.0 & \textbf{35.6} \\
\bottomrule
\end{tabular}
\caption{Experimental results of Llama-3-8B-Instruct models on the validation sets. EM: exact match; Acc: accuracy. Prompting (no tool): the non-tool-use CoT prompt (\cref{tab:prompt_no_tool_zero_shot_cot}), other prompts: the single-step tool-use prompt (\cref{tab:prompt_tool_single_step}). The $\beta$ (defined in \cref{eq:dpo}) for DPO was set to 0.1 unless specified. All numbers shown in the table are in percentages.}
\label{tab:pft_acc}
\end{table*}

\begin{table*}[t]
\small
\centering
\renewcommand{\arraystretch}{0.8}
\begin{tabular}{lcccccccccccc}
\toprule
\multirow{2}{*}{\textbf{System}}             & 
\multicolumn{3}{c}{\textbf{\triviaqa}} & \multicolumn{3}{c}{\textbf{\gsm}} & \multicolumn{3}{c}{\textbf{\nqopen}} & \multicolumn{3}{c}{\textbf{\popqa}} \\
\cmidrule(lr){2-4}\cmidrule(lr){5-7}\cmidrule(lr){8-10}\cmidrule(lr){11-13}
                   & \textbf{IR}   & \textbf{PR}   & \textbf{AR}    & \textbf{IR}  & \textbf{PR}   & \textbf{AR}    & \textbf{IR}   & \textbf{PR}   & \textbf{AR}    & \textbf{IR} & \textbf{PR} & \textbf{AR} \\
\midrule
Prompting (tools)  & 14.5 & 99.3 & 55.2  & 7.2 & 72.3 &	12.8  & 16.5 & 98.8 & 41.2  & 36.2 & 99.4 & 37.8  \\
\midrule
SFT (tools data)   & 67.0 & 99.3 & 64.6  & 75.2& 68.1 & 20.4  & 62.1 & 98.9 & 44.9  & 84.8 & 99.8 & 37.9  \\
SFT (mixture data) & 6.6  & 98.5 & 54.5  & 5.1 & 72.7 &	21.2  & 6.4  & 98.4 & 45.3  & 34.7 & 99.4 & 37.5  \\
SFT (no tool data) & 1.5  & 100.0& 53.3  & 0.5 & 100.0& 0.0   & 1.8  & 100.0& 44.4  & 9.6  & 100.0& 36.5  \\
\midrule
DPO (conversation) & 15.7 & 99.4 & 55.4  & 7.8 & 62.7 & 15.7  & 19.2 & 100.0& 50.5  & 33.8 & 99.7 & 37.6  \\
SFT + DPO          & 15.1 & 99.3 & 57.6  & 9.5 & 74.2 & 19.4  & 18.5 & 100.0& 46.5  & 33.5 & 99.7 & 37.0  \\
\midrule
DPO (response)     & 16.3 & 98.8 & 61.3  & 8.9 & 79.3 & 17.2  & 17.3 & 99.4 & 45.1  & 38.9 & 99.7 & 38.6  \\
DPO $\beta=0.01$   & 18.1 & 97.8 & 59.7  & 10.6& 72.5 &	17.4  & 20.4 & 100.0& 46.6  & 36.9 & 99.7 & 41.7  \\   
DPO $\beta=0.5$    & 16.9 & 100.0& 58.0  & 9.7 & 63.5 &	11.1  & 16.9 & 100.0& 41.4  & 37.0 & 99.5 & 40.8  \\
\bottomrule
\end{tabular}
\caption{Experimental results of Llama-3-8B-Instruct models on the validation sets. IR: invoke rate; PR: pass rate; AR: answerable rate. All prompts are the single-step tool-use prompt (\cref{tab:prompt_tool_single_step}). The $\beta$ (defined in \cref{eq:dpo}) for DPO was set to 0.1 unless specified. All numbers shown in the table are in percentages.}
\label{tab:pft_rate}
\end{table*}

\begin{table*}[t]
\centering
\small
\begin{tabular}{lcccccccc}
\toprule
\multirow{2}{*}{\textbf{System}}             & \multicolumn{2}{c}{\textbf{\triviaqa}} & \multicolumn{2}{c}{\textbf{\gsm}} & \multicolumn{2}{c}{\textbf{\nqopen}} & \multicolumn{2}{c}{\textbf{\popqa}}\\
\cmidrule(lr){2-3}\cmidrule(lr){4-5}\cmidrule(lr){6-7}\cmidrule(lr){8-9}
                   & \textbf{EM}     & \textbf{Acc}           & \textbf{EM}   & \textbf{Acc}           & \textbf{EM}    & \textbf{Acc}         & \textbf{EM}   & \textbf{Acc} \\
\midrule
Prompting (no tool)& 52.2	& 77.5          & 6.9  & 63.2 & 10.5  &41.1& 17.1 & 31.6 \\
\midrule
Prompting (tools)  & 56.0	& 75.0          & 17.8 & 58.4          & 13.8  & 38.3        & 21.2 & 33.1 \\
SFT (mixture data) & 56.0   & \textbf{78.9} & 14.9 & \textbf{61.3} & 12.2   &\textbf{40.6}& 20.3 & \textbf{35.3} \\
PFT (tools data)   & 54.8	& 73.9          & 17.3 & 57.1          & 12.3  & 37.8        & 20.2 & 33.5 \\
\bottomrule
\end{tabular}
\caption{Experimental results of Llama-3-8B-Instruct models on the test sets. EM: exact match; Acc: accuracy. Prompting (no tool): the non-tool-use CoT prompt (\cref{tab:prompt_no_tool_zero_shot_cot}), other prompts: the single-step tool-use prompt (\cref{tab:prompt_tool_single_step}). All numbers shown in the table are in percentages.}
\label{tab:test_acc}
\end{table*}

\begin{table*}[!t]
\centering
\small
\begin{tabular}{lcccccccccccc}
\toprule
\multirow{2}{*}{\textbf{System}}             & \multicolumn{3}{c}{\textbf{\triviaqa}} & \multicolumn{3}{c}{\textbf{\gsm}} & \multicolumn{3}{c}{\textbf{\nqopen}} & \multicolumn{3}{c}{\textbf{\popqa}}\\
\cmidrule(lr){2-4}\cmidrule(lr){5-7}\cmidrule(lr){8-10}\cmidrule(lr){11-13}
                   & \textbf{IR}   & \textbf{PR}   & \textbf{AR}    & \textbf{IR}   & \textbf{PR}   & \textbf{AR}    & \textbf{IR}   & \textbf{PR}   & \textbf{AR}    & \textbf{IR} & \textbf{PR}   & \textbf{AR} \\
\midrule
Prompting (tools)  & 14.5 & 98.6 & 53.1  & 10.3 & 69.6 & 13.0  & 16.9 & 100.0& 38.5  & 37.0 & 98.1 & 35.4 \\
SFT (mixture data) & 7.3  & 97.3 & 50.7  & 4.6  & 64.5 & 25.8  & 7.9  & 98.7 & 39.2  & 35.5 & 98.9 & 30.1 \\
PFT (tools data)   & 17.3 & 98.3 & 57.8  & 11.2 & 65.3 & 17.3  & 18.5 & 100.0& 41.1  & 38.0 & 98.9 & 32.4 \\
\bottomrule
\end{tabular}
\caption{Experimental results of Llama-3-8B-Instruct models on the test sets. IR: invoke rate; PR: pass rate; AR: answerable rate. All prompts are the single-step tool-use prompt (\cref{tab:prompt_tool_single_step}). All numbers shown in the table are in percentages.}
\label{tab:test_rate}
\end{table*}

\subsection{Supervised Fine-Tuning}
\label{sec:exp-sft}

\paragraph{SFT on Tools Data.} 
For the model trained on tool-use data, the results in \cref{tab:pft_acc} show that the model performance on \triviaqa, \gsm and \nqopen has degraded. The results in \cref{tab:pft_rate} show that the tool invoke rates increased significantly after SFT, indicating that training on tool-use data made the model learn to use tools intensively. While SFT teaches the LLM the behaviour of using tools, simply using these tools does not directly correlate with improved answers. This could be because (i) misusing tools may hurt model performance; (ii) the LLM might already have the essential knowledge to answer most of the questions from these two datasets, as these are typically composed of popular knowledge that could have been learned from the pre-training process. This assumption can be supported by the strong performance of the model given the prompt that no tool is allowed from \cref{tab:pft_acc}. Also, based on the results from \gsm, where no tool-use was involved, we observe the LLM can perform simple mathematical calculations, though not entirely error-free. As a result, the helpfulness of the provided tools can be limited in \triviaqa, \gsm and \nqopen. However, the results on \popqa, which contains questions involving long-tail knowledge, showed a consistent increase after using tools and fine-tuning the model on tool-use data, suggesting that the model is likely to benefit from tools when questions are less likely to be answered by its internal knowledge alone. 

\paragraph{SFT on Mixture Data.} 
For the model trained on the mixture of tool-use and non-tool-use data, the results in \cref{tab:pft_acc} show the model outperformed the one fine-tuned on tool-use data in all datasets in terms of accuracy. The model also showed a minor accuracy improvement over the not fine-tuned system. However, the tool usage has significantly declined in \triviaqa, \gsm and \nqopen, and slightly decreased in the \popqa dataset. To determine whether the performance improvement was due to enhanced tool-use ability or simply training on a larger dataset, we conducted an additional experiment using only no tool data for training. We can observe that the accuracy of models trained with tool data on \popqa is higher than the model of SFT on no tool data, suggesting the improvement stems from better tool-use ability, likely because the model has learned when to use tools.

\subsection{Preference Fine-Tuning}
\label{sec:exp-pft}

\paragraph{PFT on tool-use Data with Conversation Format.}
From the experimental results of the model optimised with DPO on conversation format tool-use data, we can observe a performance decline in the \triviaqa, \gsm and \nqopen, and a marginal accuracy improvement in \popqa, which shows a pattern similar to the model trained with SFT on tool-use data: When we used DPO with the SFT model trained on the mixture of tool-use and non-tool-use data as the base model, the results did not improve in all datasets when we compare to basing DPO on the instruction fine-tuned model. 

\paragraph{PFT on tool-use Data with Single Response Format.}
We can observe that PFT on a single response yields better results in terms of accuracy compared to the model fine-tuned on the full conversation in \triviaqa, \gsm and \popqa. This could be attributed to the fact that the DPO loss calculated on a whole conversation includes responses from tools that are redundant in the loss calculation, as we want to optimise the LLM's behaviour. Setting the preference to a single response is also suboptimal, as this did not allow the model to learn how to answer a question based on the tool responses. Therefore, how to optimise the model in the multi-turn dialogues with PFT remains an open research question. We also conducted an ablation study on the hyperparameter $\beta$ by exploring different values, showing that the choice of $\beta$ slightly impacts model accuracy.

\subsection{Results on the Test Sets}

We evaluated the proposed approaches of teaching LLMs to use tools on the test sets of \triviaqa, \gsm, \nqopen and \popqa, and the experimental results are shown in \cref{tab:test_acc} and \cref{tab:test_rate}. The SFT system was fine-tuned on the mixture of tool-use and non-tool-use data, and the PFT system was fine-tuned with tool-use data. For the PFT system, the data was constructed with the single response setting and trained with $\beta=0.5$ for DPO. The experimental results mostly showed a similar pattern to those on the validation sets, where the accuracy of tool-use systems on the \triviaqa, \gsm, and \nqopen was lower than the non-tool-use system, and the accuracy on the \popqa was better, except for the SFT system. Among the methods of teaching LLMs to use tools, the model trained with SFT showed the best accuracy, suggesting that SFT is a reasonable method to approach the task when using a limited number and variety of tools. The model trained with PFT showed better accuracy in \popqa than the prompting-based system under our experimental setting, showing the potential of using PFT to teach LLMs to use tools.

\section{Related work}

\paragraph{Prompting LLMs to Use Tools.} 
One line of research focused on investigating prompting-based methods to teach LLMs to use tools by providing tool documentation \citep{DBLP:journals/corr/abs-2308-00675} or tool descriptions and few-shot examples, \eg ReAct \citep{DBLP:conf/iclr/YaoZYDSN023}, Chameleon \cite{DBLP:conf/nips/LuPCGCWZG23}, HuggingGPT \citep{DBLP:conf/nips/0001ST00Z23}, etc. In these work, large-scale models, such as PaLM-540B \citep{DBLP:journals/jmlr/ChowdheryNDBMRBCSGSSTMRBTSPRDHPBAI23} and ChatGPT \citep{DBLP:journals/corr/abs-2303-08774}, were prompted to use tools. These studies suggested the feasibility and benefits of integrating LLMs with external tools. However, a gap remains in exploring whether a smaller model can effectively learn to use tools from prompting. Compared to prior work, our work evaluated the effectiveness of prompting LLMs across different scales to use tools. 

\paragraph{SFT for Tool Learning.} 
Another line of research applied fine-tuning-based methods to teach smaller models to use tools with curated tool-use datasets. Toolformer \citep{DBLP:conf/nips/SchickDDRLHZCS23} utilised the few-shot in-context learning ability of LLMs to generate tool-use datasets by sampling on the pre-training data and then applied data filtering. In other work where pre-training data of LLMs were inaccessible, they mainly employed more advanced LLMs, such as ChatGPT, as a teacher model to synthesise tool-use datasets and conducted supervised fine-tuning on the collected datasets (\eg ToolLLaMA \citep{DBLP:conf/iclr/QinLYZYLLCTQZHT24}, Gorilla \citep{DBLP:journals/corr/abs-2305-15334}, GPT4Tools \cite{DBLP:conf/nips/YangSLZGLS23}, inter alia). In contrast, our work began with zero-shot prompting and then leveraged tool-use datasets generated by the model itself, thereby alleviating the need for accessing tool-use examples. 

\paragraph{RLHF and Tool Learning.} 
The intersection between RLHF and tool learning is a promising yet under-explored area. TARM \cite{DBLP:conf/iclr/0040CW0T0024} showed augmenting the Reward Model (RM) in RLHF with tools enhances the agreement of RM and human judgement. TRICE \cite{DBLP:conf/naacl/QiaoGLJC024} leveraged tool execution feedback with reinforcement learning for tool learning to mitigate the problem of tool misuse adversely influencing model performance. However, an advanced LLM was still employed to synthesise tool-use datasets. Some concurrent work explored applying preference fine-tuning methods, \eg DPO and its variant, on learning to use tools to improve mathematical reasoning ability of LLMs \citep{xiong2024buildingmathagentsmultiturn, DBLP:conf/acl/WangLL24}, showcasing the benefit of utilising preference to guide model behaviour. Our work differs from these works in two aspects: (i) our work alleviates the reliance on tool-use datasets synthesised from advanced LLMs; (ii) our work explores a more comprehensive fine-tuning framework for tool learning across a broader range of tasks.

\section{Conclusion}
In this work, we studied methods for teaching LLMs to use tools without demonstrations. First, we explored teaching LLMs to use tools solely from instruction. Then, we proposed a self-training approach to synthesise datasets containing tool-use traces by instructing the LLM to use tools on two existing QA datasets, \ie \triviaqa and \gsm, and applying filtering strategies to improve data quality based on the final output and additional criteria. We then investigated methods to improve LLM's tool-use ability by fine-tuning the model with the synthetic datasets. Starting from the standard SFT objective, we then studied an under-explored approach for teaching LLMs to use tools: PFT. Experimental results suggest that proposed approaches are feasible for teaching LLMs to use tools. However, while tool-use enhances the performance of LLMs on a long-tail QA dataset, \ie \popqa, it leads to mixed results on other datasets, \ie \triviaqa, \gsm and \nqopen.

\section*{Limitations}
This work has several limitations. First, we employed a limited number of tools. Although we believe the selected tools are representative of real-world applications, as they have a relatively large action space, the generalisation of tools across various domains remains a significant research topic, which could be investigated in future work by including a broader range of tools. Second, the current tool-use dataset size in our experiments is relatively small. Although the results show the potential of using PFT to teach LLMs to use tools, future work could benefit from exploring a larger and more diverse training set, better ways of constructing training data and better loss estimation methods to fully release the power of DPO and further verify its effectiveness. Third, the self-training method, \ie using the data generated by the models themselves to improve them, typically contains multiple iterations. While we only experimented with the first iteration, future work could potentially benefit from the multiple-iteration setting.

\section*{Ethics Statement}
This work generally does not raise ethical concerns. The proposed approaches for augmenting LLMs with external tools could potentially reduce the risk of LLMs generating inaccurate information. However, there remains a possibility of potential misuse by malicious individuals using this method to enable LLMs to interact with tools for harmful purposes.

\section*{Acknowledgement}
Aryo Pradipta Gema was supported by the United Kingdom Research and Innovation (grant EP/S02431X/1), UKRI Centre for Doctoral Training in Biomedical AI at the University of Edinburgh, School of Informatics.
Xuanli He was funded by an industry grant from Cisco.
Emile van Krieken was funded by ELIAI (The Edinburgh Laboratory for Integrated Artificial Intelligence), EPSRC (grant no. EP/W002876/1).
Pietro Lesci was funded by the European Research Council (ERC) under the European Union’s Horizon 2020 Research and Innovation programme grant AVeriTeC (Grant agreement No. 865958).
Pasquale Minervini was partially funded by ELIAI, an industry grant from Cisco, and a donation from Accenture LLP.
This work was supported by the Edinburgh Compute and Data Facility (ECDF), the Edinburgh International Data Facility (EIDF) and the Data-Driven Innovation Programme at the University of Edinburgh.

\bibliography{reference}

\appendix
\section{Dataset Statistics}
\label{sec:dataset_statistic}

The dataset statistics are shown in \cref{tab:dataset_statistic}.

\begin{table*}[t!]
    \centering
    \footnotesize
    \begin{tabular}{llccc}
        \toprule
        Dataset  & Type                           & Training Set   & Validation Set & Test Set \\
        \midrule
        \multicolumn{5}{l}{\textit{Open-domain QA}} \\ 
        \triviaqa & trivia knowledge questions     & 10,000        & 1,000          & 1,000    \\
        \nqopen  & real users questions           & -             & 1,000          & 1,000    \\
        \popqa    & long-tail Wikipedia knowledge questions & -    & 1,000          & 1,000    \\
        \midrule
        \multicolumn{5}{l}{\textit{Mathematical reasoning}} \\ 
        \gsm    & grade school math questions    & 7,473         & 650            & 669      \\
        \bottomrule
    \end{tabular}
    \caption{The statistics of the datasets used in experiments.}
    \label{tab:dataset_statistic}
\end{table*}

\section{Training Details}
\label{sec:training_details}

\paragraph{Supervised Fine-Tuning.}
During experiments, the LLM was trained for 3 epochs over the curated dataset with the auto-regressive language modelling objective. We utilised Low-rank Adaptation~\citep[LoRA;][]{DBLP:conf/iclr/HuSWALWWC22}, which is one type of the Parameter-Efficient Fine-Tuning (PEFT) method, as fine-tuning the full parameters of LLMs would be expensive and time-consuming. LoRA accelerates model training by adapting the low-rank decomposition to leverage the burden of updating full parameters to update two trainable low-rank matrices instead. For LoRA training, the hyperparameter $r$ was set to 16, and the target fine-tuning modules were set to \texttt{q\_proj} and \texttt{v\_proj}, which are the default settings. As a result, around 0.08\% parameters out of the total parameters were trained. The optimiser was AdamW \citep{DBLP:conf/iclr/LoshchilovH19}. The training loss was computed on the completion only, \ie the non-LLM response messages, such as tool responses, were disregarded during loss calculation. During model training, the maximum sequence length was set to 8192 tokens. The batch size for the model was set to 4, with gradient accumulation steps of 4, so the effective batch size was 16. We also used gradient checkpointing and FlashAttention-2 \citep{DBLP:conf/iclr/Dao24} for more memory-efficient model training. All experiments were conducted on 2 A100 GPUs.

\paragraph{Preference Fine-Tuning.} 
The model was trained for 3 epochs over the curated dataset with the DPO objective. We employed DPO instead of RLHF, as the DPO pipeline is more straightforward and requires fewer computation resources. The same optimiser, LoRA setting, maximum sequence length restriction, and memory-efficient tricks as in the SFT experiments were used for model training. The batch size was set to 1 with gradient accumulation steps of 16, leading to an effective batch size of 16, which is also the same as in the SFT experiments. The maximum prompt length was set to 128, which was the default setting. All experiments were conducted on 2 A100 GPUs.

\section{Inference Details}
\label{sec:interaction}
We experimented with the instruction fine-tuned Llama 3 models with 8B\footnote{\href{https://huggingface.co/meta-llama/Meta-Llama-3-8B-Instruct}{\makesf{huggingface.co/meta-llama/Meta-Llama-3-8B-Instruct}}.} and 70B\footnote{\href{https://huggingface.co/meta-llama/Meta-Llama-3-70B-Instruct}{\makesf{huggingface.co/meta-llama/Meta-Llama-3-70B-Instruct}}.} parameters. For LLM inference, we used a default decoding setting: batch size 1, Nucleus Sampling method \citep{DBLP:conf/iclr/HoltzmanBDFC20} with a temperature of 0.6 and \texttt{top\_p} of 0.9. The maximum generated token length was set to 512. For the 70B model, we truncated the conversation length to a maximum of 8192 tokens for computational efficiency and kept other inference settings the same as the 8B model. 

The conversation list fed to LLM was in a standard chat format, composing three role components: system, user, and assistant. If tool-use is allowed, the system message was the prompt suggesting the task description, the assistant answer format, and tool lists with short descriptions if tools are allowed. The LLM then decided whether to use tools, which tools to use, and what arguments to pass. The LLM's response can contain the calling of one tool or a sequence of multiple tools. We applied a post-processing function on the LLM's response to parse it and extract tool usages with a regular expression. If tool-use information is detected, the corresponding tools will be executed with the arguments written by the LLM. Next, the tool responses, along with the past conversation history, were fed back into the LLM to generate the subsequent response. We experimented with two types of tool-use scenarios: single-step tool-use and multi-step tool-use. In the single-step tool-use scenario, the LLM can only get tool responses once but can ask for multiple tool calls simultaneously. In the multi-step tool-use scenario, the LLM can receive tool responses as many times as it wants.

\section{Evaluation Metrics}
\label{sec:metrics}

\subsection{Measuring tool-use Ability}
\label{subsec:rate}

\textbf{Invoke Rate:} This is a rate to measure the frequency of LLMs calling external tools when responding. We calculated this metric based on the tool usage amount in the first response from LLMs. The metric is defined as: 
\begin{equation}
    \text{invoke\_rate} = \frac{\text{\#tool\_usage}}{\text{\#response}} \times 100.
\end{equation}

\textbf{Pass Rate:} This is a rate to measure the percentage of LLMs successfully executing the tools, regardless of whether the content of the tool responses was related to the current conversation. We considered all tool-use instances that returned non-error responses as successful tool usage. The metric is defined as:
\begin{equation}
    \text{pass\_rate} =  \frac{\text{\#pass}}{\text{\#tool\_usage}} \times 100.
\end{equation}

\textbf{Answerable Rate:} This is a rate to measure the percentage of tool responses containing the ground truth answers, \ie determine whether the question is answerable based on the tool responses. The metric can be partial, as the LLM may employ tools to conduct intermediate steps during the question-solving process. The metric is defined as:
\begin{equation}
    \text{answerable\_rate} = \frac{\text{\#answerable}}{\text{\#tool\_usage}} \times 100.
\end{equation}

\subsection{Measuring Generation Quality}
\label{subsec:acc}

\textbf{Exact Match:} This is an accuracy commonly adopted in the realm of QA, evaluating whether the model answer is exactly the same as one of the ground truth answers. Following the normalisation process provided in the \triviaqa codebase \citep{DBLP:conf/acl/JoshiCWZ17}, the answers were normalised by removing underscores, converting into lowercase characters, removing punctuations, removing articles and then removing extra whitespaces. We applied the same normalisation process to answers of all datasets. When the normalised model answer matches one of the normalised answers from the ground truth answer list, the Exact Match score ($\text{Score}_{\text{EM}}$) of the sample equals 1; otherwise, 0. Then, we computed the Exact Match ($\text{EM}$) based on the following equation:
\begin{equation}
    \text{EM} = \left(\frac{\sum_{i=1}^{N} \text{Score}_{\text{EM}, i}}{N}\right) \times 100,
\end{equation}
where $N$ is the sample amount, and $\text{Score}_{\text{EM}, i}$ is the EM score for the $i$-th sample in the dataset.

\textbf{Accuracy:} This is an accuracy to comparing the model answer and ground truth answers. We computed the Accuracy in slightly different manners given different task types to match task scenarios better. The details are as follows:
\begin{itemize}[leftmargin=*]
    \item \textbf{Open-domain QA:} For open-domain QA datasets, \ie \triviaqa, \nqopen and \popqa, the accuracy is calculated based on the subspan score ($\text{Score}_{\text{subspan}}$), \ie check whether the model answer contains the ground truth answers, similar to \citet{DBLP:journals/tacl/LiuLHPBPL24}. The intuition is that the model answer may contain some descriptive sentences, \eg the background of the question. If the normalised model answer contains one of the normalised answers from the ground truth answer list, $\text{Score}_{\text{subspan}}$ equals 1, otherwise 0. Then, the Accuracy ($\text{Acc}_{\text{QA}}$) is computed over the whole dataset, defined as follows:
    \begin{equation}
        \text{Acc}_{\text{QA}} = \left(\frac{\sum_{i=1}^{N} \text{Score}_{\text{subspan}, i}}{N}\right) \times 100,
    \end{equation}
    where $N$ is the sample amount, and $\text{Score}_{\text{subspan}, i}$ is the subspan score for the $i$-th sample in the dataset.
    \item \textbf{Mathematical Reasoning:} For the dataset that involves mathematical computation, \ie \gsm, the accuracy is computed based on the EM score of normalised extracted answer ($\text{Score}_{\text{EM}_{\text{extract}}}$) compared to the normalised ground truth answer. The intuition is that the model answer might contain some intermediate reasoning steps that divide the questions into several sub-questions to compute the mathematical computation step-by-step in order to generate the final answer. We extracted the last digit number as the model answer, as this is the case in most cases where the model puts the answer. The Accuracy ($\text{Acc}_\text{Math}$) is computed on the whole dataset, defined as follows:
    \begin{equation}
        \text{Acc}_\text{Math} = \left(\frac{\sum_{i=1}^{N} \text{Score}_{{\text{EM}_\text{extract}}, i}}{N}\right) \times 100,
    \end{equation}
    where $N$ is the sample amount, and $\text{Score}_{{\text{EM}_\text{extract}}, i}$ is the EM score of the extract answer for the $i$-th sample in the dataset.
\end{itemize}

\section{Extended Results Analysis}
\label{sec:analysis}

\begin{figure}[t]
    \centering
    \includegraphics[width=1.0\linewidth]{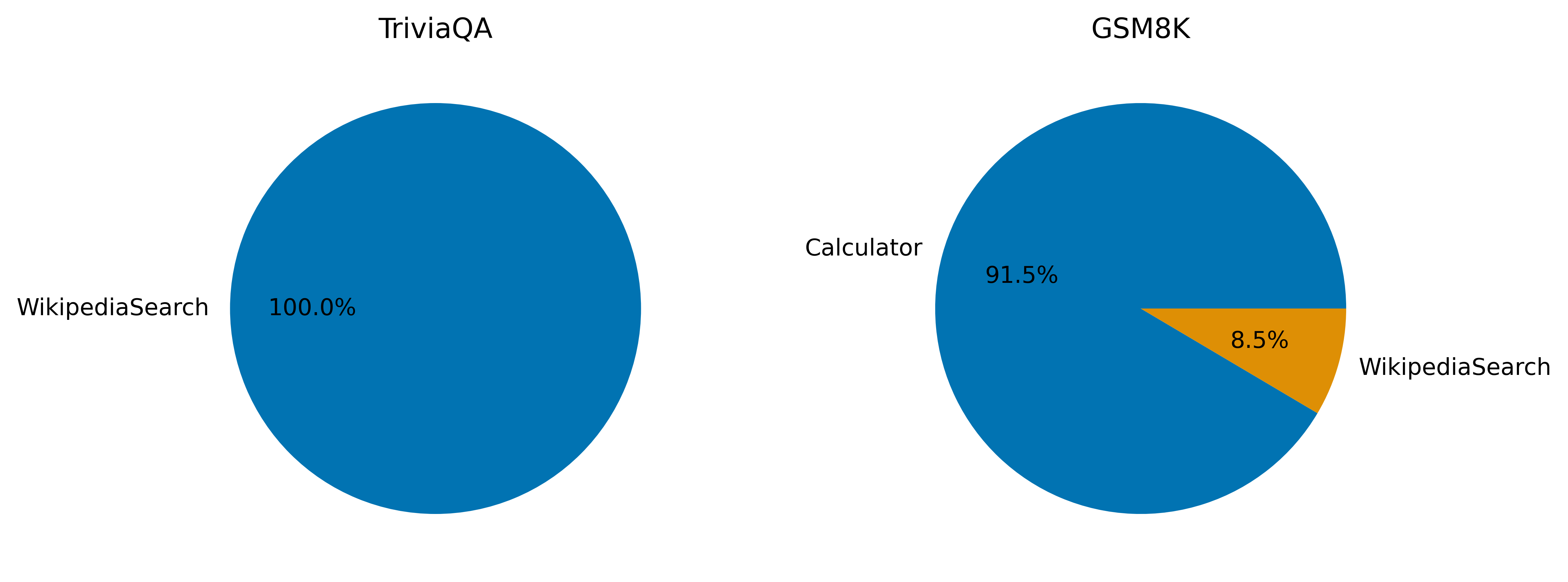}
    \caption{The distribution of tool usage by the Llama-3-8B-Instruct model on the validation sets of \triviaqa and \gsm.}
    \label{fig:tool_usage_distribution}
\end{figure}

\paragraph{Does the model understand the functionality of the provided tools?} 
We analysed the tool usage distribution on the results, shown in \cref{fig:tool_usage_distribution}. The distribution suggests that the LLM understands the functionality of given tools under our setting in most cases. Given the \triviaqa dataset, the Wikipedia search engine would be a natural choice for tools to search for trivia knowledge, and the model chose to use it at all times. In the \gsm dataset, the calculator would be the most beneficial tool. Therefore, it is also good to see that the tool usage on the dataset was dominated by it. In addition, we can observe that the model did not use the machine translator frequently, though it was provided in the tool list. This phenomenon also supports that the LLM did not randomly select tools from the provided tool list, but more likely used them based on their functionality.

\begin{table}[t]
    \centering
    \resizebox{\columnwidth}{!}{
    \begin{tabular}{lcc}
        \toprule
        Error type                           & \triviaqa     & \gsm \\
        \midrule
        No error                             & 0            & 5     \\
        Hallucination                        & \textbf{23}  & 0     \\
        Reasoning error                      & 0            & \textbf{23} \\
        Argument error                       & 0            & 1     \\
        Low-quality retrieval                & 7            & 0     \\
        Too-long context                     & 0            & 0     \\
        Infeasible actions                   & 0            & 1     \\
        Misunderstanding the tool's response & 3            & 0     \\
        \bottomrule
    \end{tabular}
    }
    \caption{The statistics of the erroneous instances randomly selected from the Llama-3-8B-Instruct model's output on the validation sets of \triviaqa and \gsm. The instance is considered erroneous when Accuracy (defined in \cref{subsec:acc}) equals 0. ``No error'' indicates that the model answer is correct, but marked incorrect due to parsing error during the answer extraction process or the ground truth answer is erroneous. }
    \label{tab:error_analysis}
\end{table}

\paragraph{Error type analysis.} 
We examined the error type of the model under both tool-use and non-tool-use settings from randomly selected erroneous instances from the validation sets of \triviaqa (30 instances) and \gsm (30 instances) on the Llama-3-8B-Instruct model, and summarised the number of common error types into \cref{tab:error_analysis}. The categories of error types were adapted and modified from \citet{DBLP:conf/nips/ZhuangYWSZ23}. We used ChatGPT \citep{openai2024chatgpt} to assist with the annotation process. For \triviaqa, we can observe that the error is dominated by ``Hallucination'', which is when the model answers with made-up facts. Also, there were several instances in which both the tools responded with low-quality information, and the model misunderstood the low-quality tool responses led to incorrect answers. The results suggest that the model would benefit from improving the tool-use ability, \ie deciding when to invoke external tools and understanding tool responses better. For \gsm, the main error type is ``Reasoning error'', in which the model performed inaccurate intermediate steps during the problem-solving process or performed incorrect calculations. Also, there were instances in which the model performed infeasible actions or passed wrong arguments to tools. The results suggest that the model could benefit from improving reasoning ability and tool-use ability.

\paragraph{Case study.} 
We examined the instances when tool-use led to inaccurate answers on the Llama-3-8B-Instruct model, and showed how low-quality retrieval and inappropriate tool-use affect the answer quality. An example from \cref{tab:use_tool_mislead} shows the LLM chose not to provide an answer when the tool response was irrelevant information, while the model provided the correct answer when not using tools. The example suggests that low-quality tool responses might adversely affect the performance of the LLM. An example in \cref{tab:use_tool_wrong} shows that the model provided a correct answer when it did not have access to tools. However, the tool response misled the reasoning process when the model accessed tools and sent an inappropriate query to the tool, although the reasoning steps were correct. It suggests that using tools might mislead the reasoning process of the model if it misuses tools.

\begin{table*}[t!]
    \centering
    \definecolor{DarkerGreen}{RGB}{34,139,34}
    \footnotesize
    \begin{tabular}{p{0.95\linewidth}}
        \toprule
        \underline{\textbf{\textsc{Non-tool-use example}}} \\
        \textbf{User:} \\
        Although both her parents are Australian, in which US state was Nicole Kidman born? \\
        \textbf{Assistant:} \\
        Thought: I will think about the birthplace of Nicole Kidman, considering she is an Australian-born actress.\\
        Answer: \textcolor{DarkerGreen}{Hawaii} \textcolor{DarkerGreen}{(correct answer)} \\
        \midrule
        \underline{\textbf{\textsc{tool-use example}}} \\
        \textbf{User:} \\
        Although both her parents are Australian, in which US state was Nicole Kidman born? \\
        \textbf{Assistant:} \\
        Thought: I don't have the information about Nicole Kidman's birthplace, so I'll use external tools to gather the information. \\
        Action: WikipediaSearch[Nicole Kidman] \\
        Rationale: I'll use the WikipediaSearch tool to find information about Nicole Kidman's birthplace. \\
        Answer: \\
        \textbf{User:} \\
        Response from tool WikipediaSearch[Nicole Kidman] are: Rabbit Hole (film) Rabbit Hole is a 2010 American tragedy film starring Nicole Kidman, Aaron Eckhart, and Dianne Wiest, and directed by John Cameron Mitchell; \dots \\
        \textbf{Assistant:} \\
        Thought: The text provided by the WikipediaSearch tool is about the movie ``Rabbit Hole'' (2010) starring Nicole Kidman, but it does not directly answer the question about Nicole Kidman's birthplace. \\
        Action: None needed, as the information provided is not relevant to the question. \\
        Rationale: The text provides information about the movie ``Rabbit Hole'', including its plot, cast, production, and reception. However, it does not mention Nicole Kidman's birthplace. \\
        Answer: \textcolor{red}{None} \textcolor{red}{(wrong answer)}\\
        \bottomrule
    \end{tabular}
    \caption{An example from the validation set of \triviaqa where a tool response misleads the Llama-3-8B-Instruct model. The latter part of the tool response is abbreviated with ``\dots'' for space efficiency.}
    \label{tab:use_tool_mislead}
\end{table*}

\begin{table*}[t!]
    \centering
    \definecolor{DarkerGreen}{RGB}{34,139,34}
    \footnotesize
    \begin{tabular}{p{0.95\linewidth}}
        \toprule
        \underline{\textbf{\textsc{Non-tool-use example}}} \\
        \textbf{User:} \\
        Zack's locker is half as big as Timothy's locker. Peter's locker is 1/4 as big as Zack's locker. If Peter's locker is 5 cubic inches, how big is Timothy's locker in cubic inches? \\
        \textbf{Assistant:} \\
        Thought: To solve this problem, I'll start by identifying the given information. Peter's locker is 5 cubic inches, and it's 1/4 as big as Zack's locker. This means Zack's locker is 5 x 4 = 20 cubic inches. Since Zack's locker is half as big as Timothy's locker, Timothy's locker is 2 x 20 = 40 cubic inches. \\
        Answer: \textcolor{DarkerGreen}{40} cubic inches. \textcolor{DarkerGreen}{(correct answer)} \\
        \midrule
        \underline{\textbf{\textsc{tool-use example}}} \\
        \textbf{User:} \\
        Zack's locker is half as big as Timothy's locker. Peter's locker is 1/4 as big as Zack's locker. If Peter's locker is 5 cubic inches, how big is Timothy's locker in cubic inches? \\
        \textbf{Assistant:} \\
        Thought: I will use external tools to solve this problem. \\
        Action: Calculator[1/4 * 5] to find the size of Zack's locker. \\
        Rationale: Since Peter's locker is 1/4 as big as Zack's locker, and we know Peter's locker is 5 cubic inches, we can multiply 5 by 4 to find the size of Zack's locker. \\
        Answer: (I will wait for the tool's response before providing the answer) \\
        \textbf{User:} \\
        Response from tool Calculator[1/4 * 5] are: 1.25 \\
        \textbf{Assistant:} \\
        Thought: Now that I have the size of Zack's locker, I will use external tools to find the size of Timothy's locker. \\
        Action: Calculator[2 * 1.25] to find the size of Timothy's locker. \\
        Rationale: Since Zack's locker is half as big as Timothy's locker, and we know Zack's locker is 1.25 cubic inches, we can multiply 1.25 by 2 to find the size of Timothy's locker. \\
        Answer: \textcolor{red}{2.5} \textcolor{red}{(wrong answer)}\\
        \bottomrule
    \end{tabular}
    \caption{An example of the Llama-3-8B-Instruct model getting the wrong answer when using tools from the validation set of \gsm.}
    \label{tab:use_tool_wrong}
\end{table*}

\section{Prompts}
\label{sec:prompts}
In addition to the single-step tool-use prompts, shown in \cref{tab:prompt_tool_single_step}, we experimented with some variations based on it for ablation. \cref{tab:prompt_no_tool_zero_shot} shows a prompt instructing the model to answer without tool-use and answer format. \cref{tab:prompt_no_tool_zero_shot_cot} shows a prompt with zero-shot CoT instructing the model to answer without tool-use, but with answer format. \cref{tab:prompt_tool_single_step_wo_rational} shows a prompt instructing the model to answer questions with single-step tool-use but without showing ``Rationale''. \cref{tab:prompt_tool_multi_step} is a multi-step variant of the single-step tool-use prompt.

\begin{table*}[h]
    \centering
    \footnotesize
    \begin{tabular}{p{0.95\linewidth}}
        \toprule
        You are an advanced AI agent designed to answer questions. Please use your own knowledge to answer the question. Answer in a few words. \\
        \bottomrule
    \end{tabular}
    \caption{System prompt for no tool-use without answer format.}
    \label{tab:prompt_no_tool_zero_shot}
\end{table*}

\begin{table*}[h]
    \centering
    \footnotesize
    \begin{tabular}{p{0.95\linewidth}}
        \toprule
        You are an advanced AI agent designed to answer questions. Please use your own knowledge to answer the question. Answer in a few words. Let's think step by step. \\ \\
        Respond in the following format: \\
        Thought: describe your thoughts on how to solve the question. \\
        Answer: provide your answer here. \\
        \bottomrule
    \end{tabular}
    \caption{System prompt for no tool-use with zero-shot CoT.}
    \label{tab:prompt_no_tool_zero_shot_cot}
\end{table*}

\begin{table*}[h]
    \centering
    \footnotesize
    \begin{tabular}{p{0.95\linewidth}}
        \toprule
        You are an advanced AI agent designed to answer questions. You can use your own knowledge to answer the questions, or use external tools to gather information before answering. However, you can only request the use of tools once. Answer in a few words. Let's think step by step. \\ \\
        Respond in the following format: \\
        Thought: describe your thoughts on how to solve the question, and decide whether to answer using your own knowledge or utilise external tools. \\
        Action: specify the tool here using the format `ToolName[query]' if you decide to use tools. \\
        Answer: (1) if using your own knowledge, provide your answer here; (2) if using tools, leave this part empty until the tool's response is received. \\ \\
        Below are the external tools you can use: \\
        1. Calculator[query]: this tool helps you perform simple mathematical computations with real numbers. Use the formula as the input query, the tool response will be the result. \\
        2. WikipediaSearch[query]: this tool helps you search for information from Wikipedia. Use a short keyword as the input query, the tool response will be the corresponding information. \\
        3. MachineTranslator[query]: this tool helps you understand low-resource languages by translating them to English. Use the sentence you want to translate as the input query, the tool response will be the translation. \\
        \bottomrule
    \end{tabular}
    \caption{System prompt for single-step tool-use without using ``Rationale'' in the response format.}
    \label{tab:prompt_tool_single_step_wo_rational}
\end{table*}

\begin{table*}[h]
    \centering
    \footnotesize
    \begin{tabular}{p{0.95\linewidth}}
        \toprule
        You are an advanced AI agent designed to answer questions. You can use your own knowledge to answer the questions, or use external tools to gather information before answering. You can request the use of tools as many times as you want. Answer in a few words. Let's think step by step. \\ \\
        Respond in the following format: \\
        Thought: decide whether to answer using your own knowledge or utilise external tools. \\
        Action: specify the tool here using the format `ToolName[query]' if you decide to use tools. \\
        Rationale: justify your answer by providing intermediate reasoning steps for your answer, based either on your own knowledge or the received tool responses. \\
        Answer: (1) if using your own knowledge, provide your answer here; (2) if using tools, leave this part empty until the tool's response is received. \\ \\
        Below are the external tools you can use: \\
        1. Calculator[query]: this tool helps you perform simple mathematical computations with real numbers. Use the formula as the input query, the tool response will be the result.  \\
        2. WikipediaSearch[query]: this tool helps you search for information from Wikipedia. Use a short keyword as the input query, the tool response will be the corresponding information. \\
        3. MachineTranslator[query]: this tool helps you understand low-resource languages by translating them to English. Use the sentence you want to translate as the input query, the tool response will be the translation. \\
        \bottomrule
    \end{tabular}
    \caption{System prompt for multi-step tool-use.}
    \label{tab:prompt_tool_multi_step}
\end{table*}

\end{document}